\documentclass[dvipsnames,format=sigconf]{acmart}
\usepackage{glossaries}
\usepackage{multirow}
\usepackage{hyperref}
\usepackage{newfloat}
\usepackage{subfigure}
\DeclareFloatingEnvironment[	% the file extension for the file used to create the list:	
fileext   = logr,% don't use log here!	% the heading for the list:	
listname  = {List of Grammars},	% the name used in captions:	
name      = Grammar,	% the default floating parameters if the environment is used	% without optional argument:	
placement = htp]{Grammar}
\AtBeginDocument{%
  \providecommand\BibTeX{{%
    \normalfont B\kern-0.5em{\scshape i\kern-0.25em b}\kern-0.8em\TeX}}}

\copyrightyear{2023}
\acmYear{2023}
\setcopyright{rightsretained}
\acmConference[GECCO '23 Companion]{Genetic and Evolutionary Computation
Conference Companion}{July 15--19, 2023}{Lisbon, Portugal}
\acmBooktitle{Genetic and Evolutionary Computation Conference Companion
(GECCO '23 Companion), July 15--19, 2023, Lisbon,
Portugal}\acmDOI{10.1145/3583133.3596399}
\acmISBN{979-8-4007-0120-7/23/07}

\begin{document}
\newacronym{snn}{SNN}{Spiking Neural Network}
\newacronym{ann}{ANN}{Artificial Neural Network}
\newacronym{dl}{DL}{Deep Learning}
\newacronym{ml}{ML}{Machine Learning}
\newacronym{ai}{AI}{Artificial Intelligence}
\newacronym{gpu}{GPU}{Graphical Processing Unit}
\title{SPENSER: Towards a NeuroEvolutionary Approach for Convolutional Spiking Neural Networks}

\author{Henrique Branquinho}
\email{branquinho@dei.uc.pt}
\orcid{1234-5678-9012}
\affiliation{%
  \institution{University of Coimbra, CISUC, DEI}
  %\streetaddress{P.O. Box 1212}
  \city{Coimbra}
  \country{Portugal}
}

\author{Nuno Lourenço}
\email{naml@dei.uc.pt}
\affiliation{%
  \institution{University of Coimbra, CISUC, DEI}
  %\streetaddress{1 Th{\o}rv{\"a}ld Circle}
  \city{Coimbra}
  \country{Portugal}}

\author{Ernesto Costa}
\email{ernesto@dei.uc.pt}
\affiliation{%
  \institution{University of Coimbra, CISUC, DEI}
  \city{Coimbra}
  \country{Portugal}
}

\renewcommand{\shortauthors}{Branquinho et al.}

\begin{abstract}

Spiking Neural Networks (SNNs) have attracted recent interest due to their energy efficiency and biological plausibility. However, the performance of SNNs still lags behind traditional Artificial Neural Networks (ANNs), as there is no consensus on the best learning algorithm for SNNs. Best-performing SNNs are based on ANN to SNN conversion or learning with spike-based backpropagation through surrogate gradients. The focus of recent research has been on developing and testing different learning strategies, with hand-tailored architectures and parameter tuning. Neuroevolution (NE), has proven successful as a way to automatically design ANNs and tune parameters, but its applications to SNNs are still at an early stage. DENSER is a NE framework for the automatic design and parametrization of ANNs, based on the principles of Genetic Algorithms (GA) and Structured Grammatical Evolution (SGE). In this paper, we propose SPENSER, a NE framework for SNN generation based on DENSER, for image classification on the MNIST and Fashion-MNIST datasets. SPENSER generates competitive performing networks with a test accuracy of 99.42\% and 91.65\% respectively. 
 
\end{abstract}

\begin{CCSXML}
<ccs2012>
   <concept>
       <concept_id>10010147.10010178.10010224</concept_id>
       <concept_desc>Computing methodologies~Computer vision</concept_desc>
       <concept_significance>500</concept_significance>
       </concept>
   <concept>
       <concept_id>10010147.10010257.10010293.10010294</concept_id>
       <concept_desc>Computing methodologies~Neural networks</concept_desc>
       <concept_significance>500</concept_significance>
       </concept>
   <concept>
       <concept_id>10003752.10003809.10003716.10011136.10011797.10011799</concept_id>
       <concept_desc>Theory of computation~Evolutionary algorithms</concept_desc>
       <concept_significance>500</concept_significance>
       </concept>
   <concept>
       <concept_id>10003752.10003766.10003771</concept_id>
       <concept_desc>Theory of computation~Grammars and context-free languages</concept_desc>
       <concept_significance>300</concept_significance>
       </concept>
 </ccs2012>
\end{CCSXML}

\ccsdesc[500]{Computing methodologies~Neural networks}
\ccsdesc[500]{Theory of computation~Evolutionary algorithms}
\ccsdesc[500]{Theory of computation~Grammars and context-free languages}
\ccsdesc[300]{Computing methodologies~Computer vision}

\keywords{spiking neural networks, neuroevolution, DENSER, computer vision}


\maketitle

\section{Introduction}

The advent of Artificial Neural Networks (ANNs) and Deep Learning (DL) has revolutionized the field of Machine Learning (ML) over the last decade, allowing for the development of high-performing models for computer vision, speech recognition, and natural language processing \cite{lecun2015deep}. However, the success of ANNs is highly dependable not only on the availability of annotated data but mostly on computationally powerful hardware such as Graphical Processing Units (GPU). This hardware dependency foresees an unsustainable future for Artificial Intelligence (AI), as the state-of-the-art models have millions of floating point parameters and require large pipelines of power-hungry hardware for training resulting in large carbon footprints \cite{patterson2021carbon}.

Spiking Neural Networks (SNNs), often called the third generation of neural networks, are biologically inspired neural network models built with spiking neurons, where information is encoded in discrete binary events over time called action potentials or spikes \cite{maass_networks_1997}. SNNs are innately sparse and highly parallelizable, which favors processing speed and energetic efficiency. Albeit still lagging behind ANNs in terms of performance, SNNs show great promise for the future of biologically plausible and sustainable AI. Current bottlenecks in SNN research include the lack of an established learning strategy, such as error backpropagation in ANNs, due to the non-differentiability of the spiking neuron's activation function, and high sensitivity to parameter tuning. Due to this, the focus of recent research, especially regarding image classification problems, has been on developing and testing different learning strategies, with hand-tailored architectures and parameter tuning usually based on successful ANN models \cite{nunes2022spiking}. However, it is unclear if these ANN architectures are suited for SNNs as well. 

Evolutionary computation (EC) methods are known to be an effective optimization tool \cite{back1993overview}, and their application to the optimization of ANNs, known as neuroevolution (NE), has proven successful both as a learning strategy as well as a way to automatically design networks and tune parameters \cite{baldominos2020automated}. DENSER \cite{assunccao2019denser, assunccao2021fast} is a NE framework for the automatic design and parametrization of ANNs, based on the principles of Genetic Algorithms (GA) and Structured Grammatical Evolution (SGE). DENSER has attained impressive results on several benchmark problems, and due to its grammar-based engine, can easily be generalized to a multitude of domains. 

In this paper, we propose SPENSER (SPiking Evolutionary Network StructurEd Representation), a NE framework for evolving Convolutional Spiking Neural Networks (CSNN) based on DENSER. This paper is a preliminary experimental study to validate SPENSER for image classification problems. In this study, we evolved the architecture and parameters of SNNs with SPENSER on the MNIST \cite{mnist} and Fashion-MNIST \cite{fashion} public datasets, using a fixed learning strategy (Backpropagation Through Time and surrogate gradients). To the best of our knowledge, this is the first work focusing on evolving SNNs trained with BPTT for image classification, including not only different architectures but different neuronal dynamics and optimizers in the search space. The main contribution of this paper is the preliminary validation of neuroevolution through SPENSER in the automatic generation of competitively performing CSNNs. The main focus of the paper is on the performance of the generated networks in terms of accuracy.

The remainder of this paper is structured as follows: Section \ref{sec:snn} provides a review of important concepts regarding SNNs; Section \ref{sec:related} covers related work regarding evolutionary approaches for SNNs; Section \ref{sec:denser} describes SPENSER; Section \ref{sec:setup} describes the experimental setup; Section \ref{experimental} analyses the experimental results, covering the evolutionary search and the testing performance of the generated models; Section \ref{sec:conclusion} provides some final remarks and suggested guidelines for future research.
\section{Spiking Neural Networks}\label{sec:snn}

Spiking Neural Networks (SNNs) are a class of neural network models built with spiking neurons where information is encoded in the timing and frequency of discrete events called spikes (or action potentials) over time  \cite{maass_networks_1997}. Spiking neurons can be characterized by a membrane potential $V(t)$ and activation threshold $V_{thresh}$. The weighted sum of inputs of the neuron increases the membrane potential over time. When the membrane potential reaches its activation threshold, a spike is generated (fired) and propagated to subsequent connections. In a feed-forward network, inputs are presented to the network in the form of spike trains (timed sequences of spikes) over $T$ time steps, during which time spikes are accumulated and propagated throughout the network up to the output neurons. 

There are a number of spiking neuron models that vary in biological plausibility and computational cost, such as the more realistic and computationally expensive Hodgkin-Huxley \cite{Hodgkin1952}, to the more simplistic and computationally lighter models such as the Izhikevich \cite{izhikevich2003simple}, Integrate-and-Fire (IF) \cite{lapicque1907recherches} and Leaky Integrate-and-Fire (LIF) \cite{dutta2017leaky}. We refer to Long and Fang \cite{long_review_2010} for an in-depth review of existing spiking neuron models and their behaviour.

The LIF neuron is the most commonly used in the literature due to its simplicity and low computational cost. The LIF neuron can be modulated as a simple parallel Resistor-Capacitor (RC) circuit with a "leaky" resistor:

\begin{equation}
    C \frac{dV}{dt} = - g_L ( V(t) - E_L ) + I(t)
\label{eq:lif}
\end{equation}

In Eq. \ref{eq:lif}, $C$ is a capacitor, $g_L$ is the "leaky" resistor (conductor), $E_L$ is the resting potential and $I(t)$ is the current source (synaptic input) that charges up the capacitor to increase the membrane potential $V(t)$. Solving this differential equation through Euler method (demonstration in \cite{eshraghian_training_2022}), we can calculate a neuron's membrane potential at a given timestep $t$ as:

\begin{equation}
    V[t] = \beta V[t-1] + W X[t] - Act[t-1] V_{thresh}
\label{eq:lif2}
\end{equation}

In Eq. \ref{eq:lif2}, $\beta$ is the decay rate of the membrane potential, $X[t]$ is the input vector (corresponding to $I(t)$), $W$ is the vector of input weights, and $Act[t]$ is the activation function. The activation function can be defined as follows:

\begin{equation}
    Act[t] = 
\left\{
    \begin{array}{lr}
        1, & \text{if } V[t] > V_{thresh}  \\
        0, & otherwise
    \end{array}
\right\}
\label{eq:act}
\end{equation}

A LIF neuron's membrane potential naturally decays to its resting state over time if no input is received ($\beta V[t-1]$). The potential increases when a spike is received from incoming connections, proportionally to the connection's weight ($W X[t]$). When the membrane potential $V(t)$ surpasses the activation threshold $V_{thresh}$ a spike is emitted and propagated to outgoing connections and the membrane's potential resets ($-Act[t-1]V_{thresh}$). Resetting the membrane's potential can be done either by subtraction, as is done in the presented example, where $V_{thresh}$ is subtracted at the onset of a spike; or to zero, where the membrane potential is set to 0 after a spike. A refractory period is usually taken into account where a neuron's potential remains at rest after spiking in spite of incoming spikes. The decay rate and threshold can be static or trainable. 

Existing frameworks such as \textit{snntorch} \cite{eshraghian_training_2022} allow for the development of SNNs by integration of spiking neuron layers in standard ANN architectures such as Convolutional Neural Networks, by simply replacing the activation layer with a spiking neuron layer.  

% \begin{figure}[b]
% \centering
    
%     \includegraphics[width=6cm]{figs/lif.png}
%     \caption{LIF Neuron}
%     \label{fig:LIF}
% \end{figure}

\subsection{Information Coding}

Spiking systems rely on discrete events to propagate information, so the question arises as to how this information is encoded. We focus on two encoding strategies: rate coding and temporal coding. In \textbf{rate coding}, information is encoded in the frequency of firing rates. This is the case in the communication between photoreceptor cells and the visual cortex, where brighter inputs generate higher frequency firing rates as opposed to darker inputs and respectively lower frequency firing rates \cite{hubel1962receptive}. ANNs rely on rate coding of information, as each neuron's output is meant to represent an average firing rate. In \textbf{temporal coding}, information is encoded in the precise timing of spikes. A photoreceptor system with temporal coding would encode a bright input as an early spike and a dark input as a last spike. When considering the output of an SNN for a classification task, the predicted class would either be: the one with the highest firing frequency, using rate coding; the one that fires first, using temporal coding. 

Temporal coding is advantageous in terms of speed and power consumption, as fewer spikes are needed to convey information, resulting in more sparse events which translate to fewer memory accesses and computation. On the other hand, rate coding is advantageous in terms of error tolerance, as the timing constraint is relaxed to the overall firing rate, and promoting learning, as the absence of spikes can lead to the "dead neuron" problem, where no learning takes place as there is no spike in the forward pass. Increased spiking activity prevents the "dead neuron" problem.

\subsection{Learning}\label{snn:learning}

Learning in SNNs remains one of the biggest challenges in the community due to the non-differentiability of the activation function of spiking neurons (Eq. \ref{eq:act}), which does not allow for the direct transposition of the error backpropagation algorithm. 

Commonly used learning strategies include unsupervised learning through Spike-Timing-Dependent Plasticity (STDP) \cite{diehl2015unsupervised}, offline conversion from trained ANNs to SNNs (also known as shadow training) \cite{rueckauer2017conversion,deng2021optimal}, and supervised learning through backpropagation either using spike times \cite{bohte2002error} or adaptations of the activation function to a continuous-valued function \cite{ledinauskas2020training,huh2018gradient,hunsberger2015spiking, neftci2019surrogate,shrestha2018slayer}. In this work, we focus on the latter, by training SNNs using backpropagation through time (BPTT) and surrogate gradients. 

BPTT is an application of the backpropagation algorithm to the unrolled computational graph over time, usually applied to Recurrent Neural Networks (RNNs) \cite{werbos1990backpropagation}. In order to bypass the non-differentiability of the spiking neuron's activation function, one can use surrogate gradients, by approximating the activation function with continuous functions centered at the activation threshold during the backward pass of backpropagation \cite{neftci2019surrogate}. 

%The activation function of a LIF neuron is effectively a Heaviside step function shifted to $V_{thresh}$ ($Act \in \{0,1\}$, Eq. \ref{eq:act}), whose derivative with respect to the membrane potential $V$ is a Dirac delta function ($\frac{\partial Act}{\partial V} \in \{0,\infty \}$ ). When a neuron does not spike ($V < V_{thresh}$), the derivative will evaluate to 0 and no learning will take place in the backward pass (the "dead neuron" problem aforementioned, see Figure \ref{fig:surrogate}). However, if we consider a continuous function centered around the activation threshold $V_{thresh}$, the derivative will evaluate to non-zero values and serve as a biased estimator of the gradient to flow backward. During the forward pass, the Heaviside operator is used to generate spikes. During the backward pass, the surrogate gradient replaces the Dirac delta to calculate the gradient. An example of applying the sigmoid function as a surrogate gradient is shown in Fig. \ref{fig:surrogate}.

% \begin{figure}[h]
%     \centering
%     \includegraphics[width=6cm]{figs/surrogate.png}
%     \caption{Surrogate gradient}
%     \label{fig:surrogate}
% \end{figure}

In this experimental study, we considered two surrogate gradient functions available in \textit{snntorch} \cite{eshraghian_training_2022}:
\begin{itemize}
    \item \textbf{Fast-Sigmoid}

    \begin{equation}
        Act \approx \frac{V}{1 + k|V|}
    \end{equation}

     \item \textbf{ATan} - Shifted arc-tan function

    \begin{equation}
        Act \approx \frac{1}{\pi} arctan(\pi V \frac{\alpha}{2})
    \end{equation}
    
\end{itemize}

Regarding the loss function, there are a number of choices available depending on the output encoding of the network (rate vs temporal), that calculate the loss based on spikes or on membrane potential. For this experimental study, we considered rate encoding for inputs and outputs, and as such, chose the \textbf{Mean Square Error Spike Count Loss} (adapted from \cite{shrestha2018slayer}). The spike counts of both correct and incorrect classes are specified as targets as a proportion of the total number of time steps (for example, the correct class should fire 80\% of the time and the incorrect classes should only fire 10\%). The target firing rates are not required to sum to 100\%. After a complete forward pass, the mean square error between the actual ($\sum_{t=0}^{T} Act[t]$) and target ($\hat{Act}$) spike counts of each class $C$ is calculated and summed together (Eq.\ref{eq:mse}).

\begin{equation}
    \mathcal{L} = \frac{1}{T} \sum_{j=0}^{C-1} (\sum_{t=0}^{T} Act_j[t] - \hat{Act}_j )^2
\label{eq:mse}
\end{equation}

\section{Related Work}\label{sec:related}

Recent works blending EC and SNNs are mostly focused on evolving a network's weights, using evolutionary approaches as a learning strategy \cite{lopez-vazquez_evolutionary_2019, kozdon_evolution_2018, lu_neuroevolution_2022}. 

Schuman et al. \cite{schuman2020evolutionary} proposed Evolutionary Optimization for Neuromorphic Systems, aiming to train spiking neural networks for classification and control tasks, to train under hardware constraints, to evolve a reservoir for a liquid state machine, and to evolve smaller networks using multi-objective optimization. However, they focus on simple machine learning classification tasks and scalability is unclear. Elbrecht and Schuman \cite{elbrecht_neuroevolution_2020} used HyperNeat \cite{stanley2009hypercube} to evolve SNNs focusing on the same classification tasks. Grammatical Evolution (GE) has also been used previously by López-Vázquez et al. \cite{lopez-vazquez_evolutionary_2019} to evolve SNNs for simple classification tasks.

The current state of the art in the automatic design of CSNN architectures are the works of Kim et al. \cite{kim_neural_2022} and
AutoSNN by Na et al. \cite{na_autosnn_2022}. Both works focus on Neural Architecture Search (NAS), with an evolutionary search component implemented in AutoSNN, and attain state-of-the-art performances in the CIFAR-10, CIFAR-100 \cite{cifar10}, and TinyImageNet datasets. However, both works fix the generated networks' hyperparameters such as LIF neuron parameters and learning optimizer. Our work differs from these works by incorporating these properties in the search space.

\section{SPENSER}\label{sec:denser}

% Deep learning has proven to be an efficient tool to generate classification and prediction models due to their inate ability to uncover patterns and filter out noise from raw data, serving as black box models with excellent results. However, the success of these models is heavily dependent on the network design and parameterization. The field of AutoML is dedicated to ease this process, allowing for the automation of machine learning pipelines design that better suit the dataset and task at hand. In the scope of AutoML applied to deep learning we have neuroevolution, which consists in employing evolutionary strategies for the generation of neural networks. One such framework is DENSER. 

\begin{figure}[t]
\centering
    
    \includegraphics[width=7.92cm]{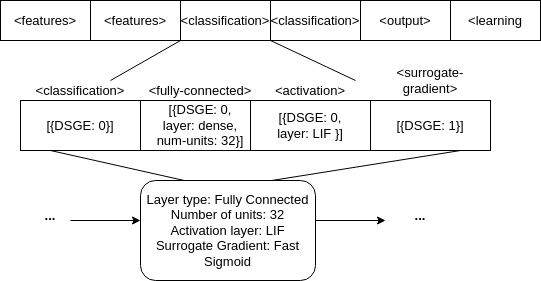}
    \caption{Individual generation by SPENSER. The first line represents the GA level where the macrostructure of the network is defined (this individual has 2 \textit{features} units and 2 \textit{classification} units). The second line represents the specification of a \textit{classification} unit through DSGE. Each number in the DSGE level represents the index of the chosen expansion rule for the current non-terminal. The last line is the resulting phenotype of the layer in question \cite{assunccao2019denser}.}
    \label{fig:denser}
\end{figure}

SPENSER (SPiking Evolutionary Network StructurEd Representation) is a general-purpose evolutionary-based framework for the automatic design of SNNs, based on DENSER \cite{assunccao2019denser,assunccao2021fast}, combining the principles of Genetic Algorithms (GA) \cite{9783540731894} and Dynamical Structured Grammatical Evolution (DSGE) \cite{lourencco2015sge,lourencco2018structured}. SPENSER works on a two-level basis, separating the GA and the DSGE level, which allows for the modeling of the overall network structure at the GA level while leaving the network layer's specifications for the DSGE (Figure \ref{fig:denser}). The use of a grammar is what makes SPENSER a general-purpose framework, as one solely needs to change the grammar to handle different network and layer types, problems and parameters range. 

The GA level encodes the macrostructure representing the sequence of evolutionary units that form the network. Each unit corresponds to a nonterminal from the grammar that is later expanded through DSGE. With this representation, we can encode not only the network's layers as evolutionary units but also the optimizer and data augmentation. Furthermore, by assigning each evolutionary unit to a grammar nonterminal, we can encode prior knowledge and bound the overall network architecture.

The DSGE level is responsible for the specification of each layer's type and parameters, working independently from the GA level. DSGE represents an individual's genotype as a set of expansion choices for each expansion rule in the grammar. Starting from a nonterminal unit from the GA level, DSGE follows the expansions set in the individual's genotype until all symbols in the phenotype are nonterminals. Rules for the layer types and parameters are represented as a Context-Free Grammar (CFG), making it easier to adapt the framework to different types of networks, layers and problem domains. 

An example encoding to build CSNNs could be defined by Grammar \ref{gram:example} and the following GA macro structure:

\begin{align*}
%\begin{displaymath}
[(features,1,10), & \,(classification,1,3),\\
(output,1,1), & \,(learning,1,1)]
%\end{displaymath}
\end{align*}

\begin{Grammar}
\centering
\scalebox{0.7}{
        \setlength{\fboxrule}{0pt}
  \fbox{%
				\parbox{0.1\textwidth}{%
					\begin{align*}
{<}\text{features}{>} ::= & \, {<}\text{convolution}{>} \text{  } {<}\text{activation}{>}| {<}\text{convolution}{>} \text{  }{<}\text{pooling}{>}\\
& \,{<}\text{activation}{>}\\
 {<}\text{convolution}{>} ::= & \, \text{layer}:\text{conv} \text{  } [\text{num-filters},\text{int},1,32,256] \text{  }
                  [\text{filter-shape},\text{int},1,2,5] \\
 {<}\text{classification}{>} ::= & \, {<}\text{fully-connected}{>} \text{  } {<}\text{activation}{>}\\
{<}\text{fully-connected}{>} ::=  & \, \text{layer}:\text{dense} \text{  }[\text{num-units},\text{int},1,16,128] \\
{<}\text{activation}{>} ::=  & \, \text{layer}:\text{LIF} \text{  }  {<}\text{surrogate-gradient}{>} \\
{<}\text{surrogate-gradient}{>} ::= & \, {<}\text{ATan}{>} \text{  } | {<}\text{FastSigmoid}{>}\\
{<}\text{output}{>} ::=  & \, \text{layer}:\text{dense} \text{  }\text{num-units}:10 \text{  } {<}\text{activation}{>}\\
{<}\text{learning}{>} ::= & \, {<}\text{Adam}{>} \text{  } | {<}\text{SGD}{>}\\
...& \,
					\end{align*}}}}

 \caption{Example of a Convolutional Spiking Neural Network grammar.}
\label{gram:example}
 \end{Grammar}

The numbers in each macro unit represent the minimum and maximum number of units that can be incorporated into the network. With this example, the $features$ block encodes layers for feature extraction, and therefore we can generate networks with convolutional and pooling layers, followed by 1 to 3 fully connected layers from the $classification$ units. The activation layers are restricted to LIF nodes with different surrogate gradient options. The $learning$ unit represents the optimizer used for learning and its parameters. The $output$ unit encodes the network's output layer. Numeric parameters are defined by their type, the number of parameters to generate, and the range of possible values. 

Regarding variation operators, SPENSER relies on mutations on both levels. At the GA level, individuals can be mutated by adding, replicating, or removing genes i.e. layers. At the DSGE level, mutation changes the layers' parameters by grammatical mutation (replacing grammatical expansions), integer mutation (replacing an integer parameter with a uniformly generated random one), and float mutation (modifying a float parameter through Gaussian perturbation). SPENSER follows a $(1 + \lambda)$ evolutionary strategy where the parent individual for the next generation is chosen by highest fitness and mutated to generate the offspring. This evolutionary strategy was chosen due to the computational demands of the network training process, which limits the population size in regard to execution time.

\section{Experimental Setup}\label{sec:setup}

For this experimental study, we evolved and tested networks on the MNIST \cite{mnist} and Fashion-MNIST \cite{fashion} datasets, available through the Torchvision library of Pytorch. All images were converted to grayscale and their original size was kept (28x28). In order to apply SNNs to these datasets, the images were converted to spike trains using rate coding. The pixel values are normalized between 0 and 1 and each pixel value is used as a probability in a Binomial distribution, which is then sampled from to generate spike trains of length $T$ time steps. No data augmentation was used. We considered different time steps for each dataset according to their complexity.

% \begin{figure}[h]
%     \centering
%     \includegraphics[width=7.92cm]{figs/ratecoding.png}
%     \caption{Rate coding}
%     \label{fig:ratecoding}
% \end{figure}

Datasets were split in three subsets: EvoTrain, Fitness and Test. The Test split is the one provided by Torchvision. The EvoTrain and Fitness splits are a 70/30 split of the original Train split. Each independent run generates different EvoTrain and Fitness splits. Table \ref{tab:datasets} summarises the chosen time steps and the number of samples per split for each dataset.

\begin{table}[ht]
\caption{Time steps and number of samples per split for each dataset (MNIST and Fashion-MNIST).}
\begin{tabular}{|l|c|lc|l|}
\hline
                       &                         & \multicolumn{2}{l|}{\textbf{Train}}                                        &                                             \\ \cline{2-5} 
                       & \textbf{Time Steps (T)} & \multicolumn{1}{l|}{EvoTrain}               & \multicolumn{1}{l|}{Fitness} & \textbf{Test}                               \\ \hline
\textbf{MNIST}         & 10                      & \multicolumn{1}{c|}{\multirow{2}{*}{42000}} & \multirow{2}{*}{18000}       & \multicolumn{1}{c|}{\multirow{2}{*}{10000}} \\
\textbf{F-MNIST} & 25                      & \multicolumn{1}{c|}{}                       &                              & \multicolumn{1}{c|}{}                       \\ \hline
\end{tabular}
\label{tab:datasets}
\end{table}

As this is a preliminary study to validate SPENSER, we settled on one-pass training of individuals as a trade-off between speed and accuracy. During the evolutionary search, individuals are trained on the EvoTrain split for 1 epoch and tested against the Fitness split for fitness assignment. After the evolutionary search is complete, the best individual is further trained for 50 epochs on the entire Train set, and tested against the Test set for accuracy assessment. 

We used \textit{snntorch} \cite{eshraghian_training_2022} to assemble, train and evaluate SNNs based on rate coding. Individuals are trained using BPTT and the chosen loss function was the Mean Square Error Spike Count described in Section \ref{snn:learning}, with a target spiking proportion of 100\% for the correct class and 0\% for the incorrect class. The predicted class for a given instance is calculated based on the highest spike count of the output neurons. Accuracy is used as the fitness metric during the evolutionary search and as the final performance assessment of the best found individuals. 

The macro structure of individuals for the GA level was set as:

\begin{align*}
%\begin{displaymath}
[(features,1,6), & \,(classification,1,4),\\
(output,1,1), & \,(learning,1,1)]
%\end{displaymath}
\end{align*}

Because we are dealing with an image recognition problem, we defined a grammar that contains primitives allowing for the construction of CSNNs, as shown in Grammar \ref{gram:cnn}. Following is a brief description of the grammar.

$features$ units can be expanded to either Convolutional + Activation, Convolutional + Pooling + Activation, or Dropout layers. Convolutional layers are defined by the number of filters, filter shape, stride, padding and bias. Pooling layers are defined by the pooling type (max or average) and the kernel size. $classification$ units can be expanded to either Fully-Connected + Activation or Dropout layers. Fully-Connected layers are defined by the number of units. Dropout layers are defined by the dropout rate. The $output$ unit is set as a Fully-Connected + Activation where the number of units is fixed to the number of classes. Activation layers are currently limited to LIF neurons. LIF neurons are defined by the decay rate $\beta$, the activation threshold $V_{thresh}$, and the reset mechanism (subtraction or zero). Furthermore, they are also defined by the surrogate gradient function, which in this case can be either the ATan or the Fast-Sigmoid functions described in Section \ref{snn:learning}. The $learning$ unit encodes the optimizer and can be expanded to either Stochastic Gradient Descent, Adam, or RMSProp. We increased the probability of choosing feature extraction layers over dropout for $features$ units (Grammar \ref{gram:cnn}, line 1).

\begin{Grammar}[]
%\centering
\scalebox{0.7}{
        \setlength{\fboxrule}{0pt}
  \fbox{%
				\parbox{0.1\textwidth}{%
					\begin{align*}
{<}\text{features}{>}::=& \,{<}\text{aux-convolution}{>}\text{ }|\text{ }{<}\text{aux-convolution}{>}\text{ }|\text{ }{<}\text{aux-convolution}{>}\\
& \,|\text{ }{<}\text{dropout}{>}\\ 
{<}\text{aux-convolution}{>}::=& \,{<}\text{convolution}{>}\text{ }{<}\text{pooling}{>}\text{ }{<}\text{activation}{>}\\ 
{<}\text{activation}{>}::=& \,\text{layer}:\text{act}\text{ }{<}\text{beta}{>}\text{ }{<}\text{threshold}{>}\text{ }{<}\text{surr-grad}{>}\text{ }{<}\\
& \,\text{reset-mechanism}{>}\\ 
{<}\text{reset-mechanism}{>}::=& \,\text{reset}:\text{subtract}\text{ }|\text{ }\text{reset}:\text{zero}\\ 
{<}\text{beta}{>}::=& \,[\text{beta},\text{float},1,0,1]\text{ }{<}\text{beta-trainable}{>}\\ 
{<}\text{threshold}{>}::=& \,[\text{threshold},\text{float},1,0.5,1.5]\text{ }{<}\text{threshold-trainable}{>}\\ 
{<}\text{beta-trainable}{>}::=& \,\text{beta-trainable}:\text{True}\text{ }|\text{ }\text{beta-trainable}:\text{False}\\ 
{<}\text{threshold-trainable}{>}::=& \,\text{threshold-trainable}:\text{True}\text{ }|\text{ }\text{threshold-trainable}:\text{False}\\ 
{<}\text{surr-grad}{>}::=& \,\text{surr-grad}:\text{atan}\text{ }|\text{ }\text{surr-grad}:\text{fast-sigmoid}\\ 
{<}\text{pooling}{>}::=& \,{<}\text{pool-type}{>}\text{ }[\text{kernel-size},\text{int},1,2,4]\text{ }|\text{ }\text{layer}:\text{no-op}\\ 
{<}\text{pool-type}{>}::=& \,\text{layer}:\text{pool-avg}\text{ }|\text{ }\text{layer}:\text{pool-max}\\ 
{<}\text{classification}{>}::=& \,{<}\text{fully-connected}{>}\text{ }{<}\text{activation}{>}\text{ }|\text{ }{<}\text{dropout}{>}\\ 
{<}\text{convolution}{>}::=& \,\text{layer}:\text{conv}\text{ }[\text{num-filters},\text{int},1,32,128]\text{ }[\text{filter-shape},\text{int},1,2,4]\text{ } \\
& \,[\text{stride},\text{int},1,1,3]\text{ }{<}\text{padding}{>}\text{ }{<}\text{bias}{>}\\ 
{<}\text{padding}{>}::=& \,\text{padding}:\text{same}\text{ }|\text{ }\text{padding}:\text{valid}\\ 
{<}\text{dropout}{>}::=& \,\text{layer}:\text{dropout}\text{ }[\text{rate},\text{float},1,0,0.5]\\ 
{<}\text{fully-connected}{>}::=& \,\text{layer}:\text{fc}\text{ }[\text{num-units},\text{int},1,32,256]\text{ }{<}\text{bias}{>}\\ 
{<}\text{bias}{>}::=& \,\text{bias}:\text{True}\text{ }|\text{ }\text{bias}:\text{False}\\ 
{<}\text{output}{>}::=& \,{<}\text{fully-last}{>}\text{ }{<}\text{activation}{>}\\ 
{<}\text{fully-last}{>}::=& \,\text{layer}:\text{fc}\text{ }\text{num-units}:10\text{ }\text{bias}:\text{True}\\ 
{<}\text{learning}{>}::=& \,{<}\text{gradient-descent}{>}\text{ }|\text{ }{<}\text{rmsprop}{>}\text{ }|\text{ }{<}\text{adam}{>}\text{ }\\ 
{<}\text{gradient-descent}{>}::=& \,\text{learning}:\text{gradient-descent}\text{ }[\text{lr},\text{float},1,0.0001,0.1]\text{ } \\
& \,[\text{momentum},\text{float},1,0.68,0.99]\text{ }[\text{decay},\text{float},1,0.000001,0.001]\\  
& \,{<}\text{nesterov}{>}\\ 
{<}\text{nesterov}{>}::=& \,\text{nesterov}:\text{True}\text{ }|\text{ }\text{nesterov}:\text{False}\\ 
{<}\text{adam}{>}::=& \,\text{learning}:\text{adam}\text{ }[\text{lr},\text{float},1,0.0001,0.1]\\
& \,[\text{beta}1,\text{float},1,0.5,0.9999] [\text{beta}2,\text{float},1,0.5,0.9999]\\
& \,[\text{decay},\text{float},1,0.000001,0.001]\text{ }{<}\text{amsgrad}{>}\\ 
{<}\text{amsgrad}{>}::=& \,\text{amsgrad}:\text{True}\text{ }|\text{ }\text{amsgrad}:\text{False}\\ 
{<}\text{rmsprop}{>}::=& \,\text{learning}:\text{rmsprop}\text{ }[\text{lr},\text{float},1,0.0001,0.1]\text{ }[\text{rho},\text{float},1,0.5,1]\\
& \,[\text{decay},\text{float},1,0.000001,0.001]
					\end{align*}}}}

 \caption{Convolutional Spiking Neural Network grammar.}
\label{gram:cnn}
 \end{Grammar}

Regarding SPENSER's main hyper-parameters, we followed the recommendations of \cite{assunccao2019denser,assunccao2021fast}, summarised in Table \ref{tab:denser_params}. The table is divided in two parts: i) evolutionary parameters, specifying the evolutionary engine properties such as number of generations, number of parents ($\mu$), number of offspring ($\lambda$), mutation rates and fitness function; ii) training parameters, specifying the overall learning parameters fixed for all networks.

\begin{table}[htp]
\centering
\caption{Hyper-parameters for SPENSER.}
\begin{tabular}{lll}
\textbf{Evolutionary Parameter} & \textbf{Value}            &  \\
Number of runs                  & 5                       &  \\
Number of Generations           & 200                       &  \\
$\mu$ (\#Parents)                   & 1                         &  \\
$\lambda$ (\#Offspring)                 & 10                        &  \\
Add Layer Rate                  & 25\%                      &  \\
Duplicate Layer Rate            & 15\%                      &  \\
Remove Layer Rate               & 25\%                      &  \\
Layer DSGE Rate                & 15\%                      &  \\
Learning DSGE Rate & 30\% & \\
Gaussian Perturbations & (0 , 0.15) & \\
Fitness Function                & Accuracy                  &  \\
                                &                           &  \\
\textbf{Training Parameters}    & \textbf{Value}            &  \\
Number of epochs              & 1                &  \\
Batch Size & 64 & \\
Loss Function                   &  Mean Square Error Spike Count & \\
Correct Rate  & 1.0 & \\
Incorrect Rate & 0.0 & \\
\end{tabular}

\label{tab:denser_params}
\end{table}

All the code, configuration files, grammar, and execution instructions for these experiments are publicly available at GitHub \footnote{\url{https://github.com/henriquejsb/spenser}}.
\section{Experimental Results}\label{experimental}

\subsection{Evolutionary Search}

\begin{figure}[htp]
\centering
\subfigure[Best Fitness]{
    \includegraphics[width=.31\textwidth]{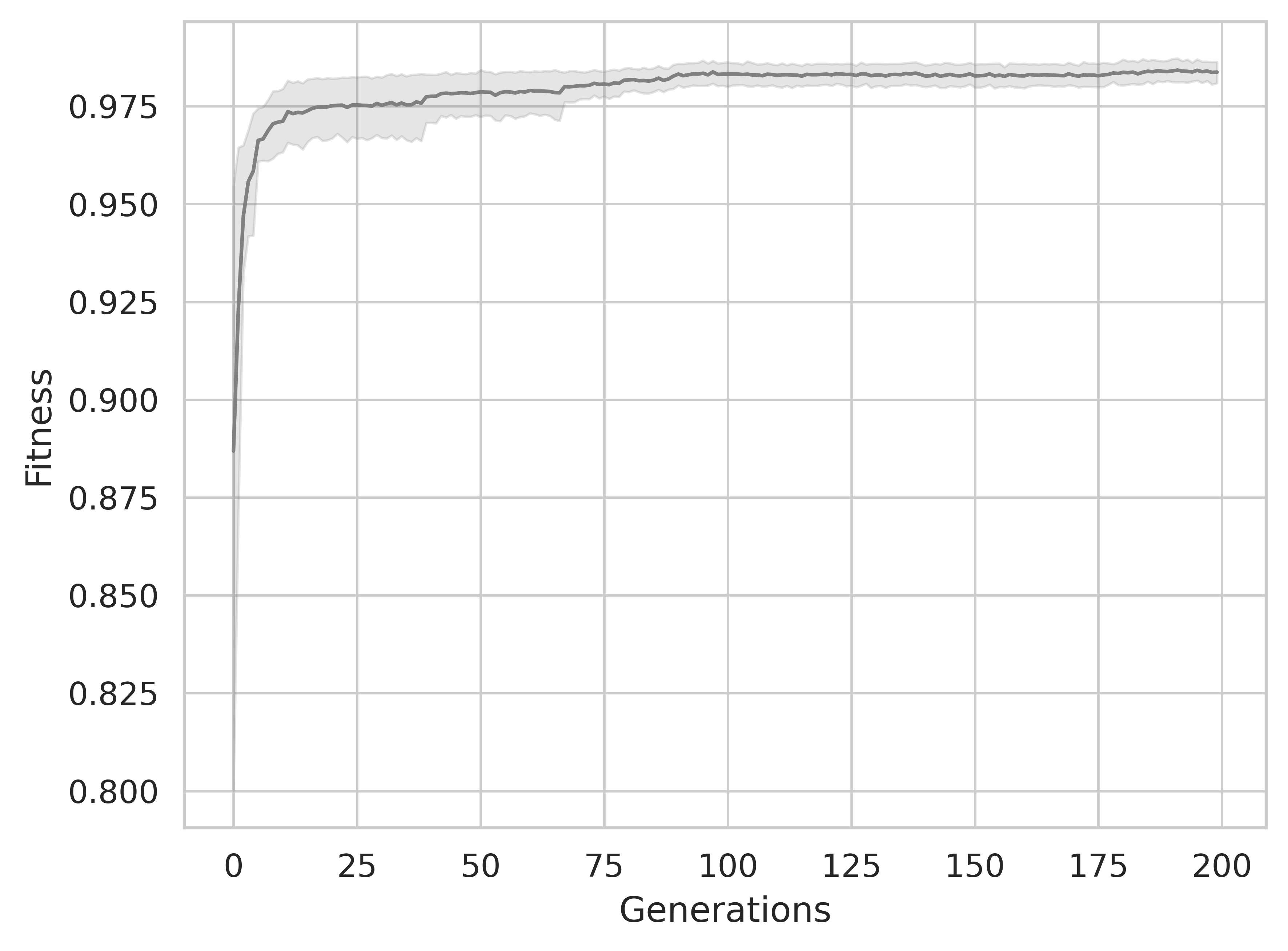}
    \label{fig:mnist_best}
    
}
\subfigure[Average Fitness]{
    \includegraphics[width=.31\textwidth]{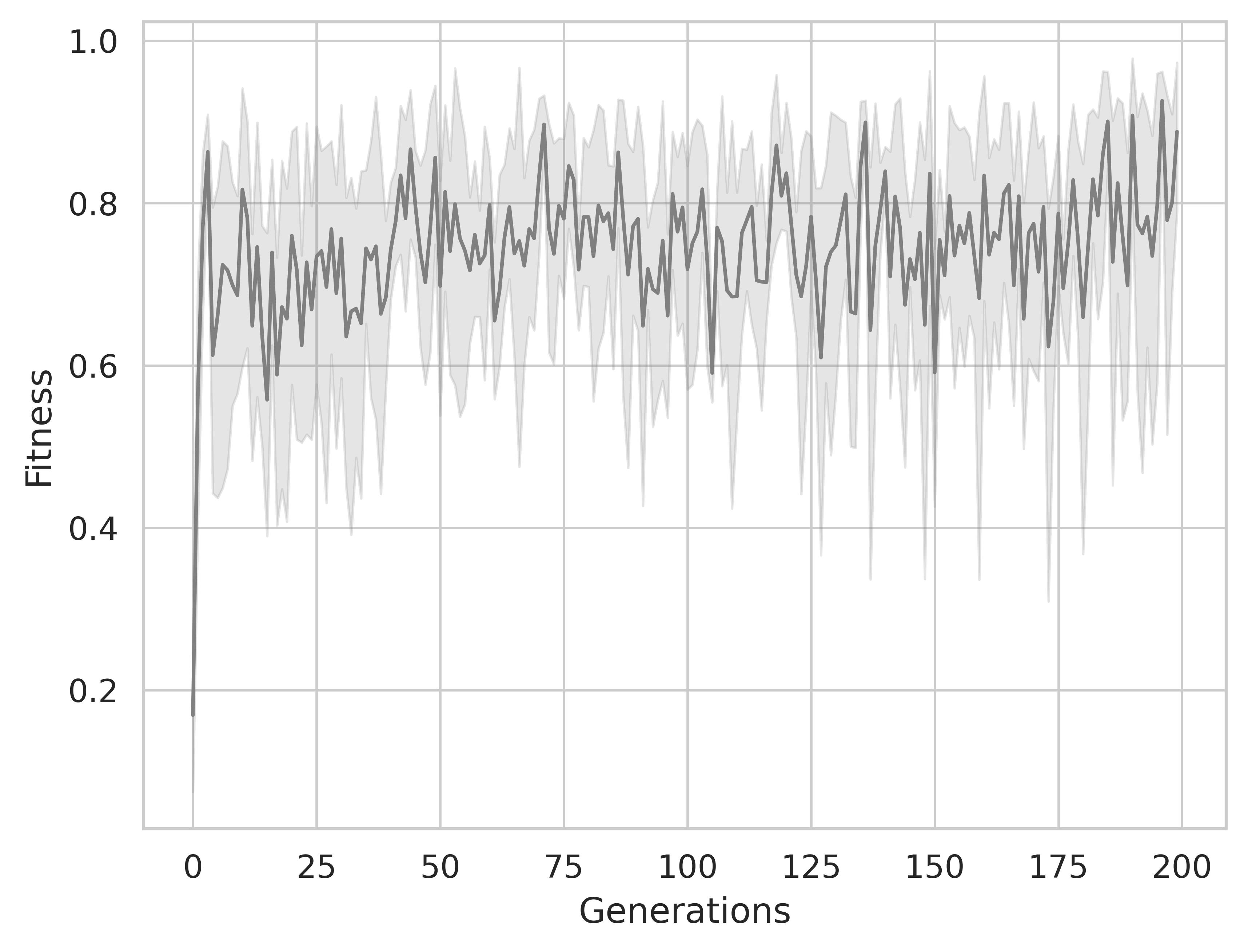}
    \label{fig:mnist_average}
}
%\\[\smallskipamount]
% \subfigure[New Best Individuals]{
%     \includegraphics[width=.31\textwidth]{figs/mnist_new_parents.png}
%     \label{fig:mnist_new}
% }
\subfigure[Fitness of best found individuals]{
    \includegraphics[width=.31\textwidth]{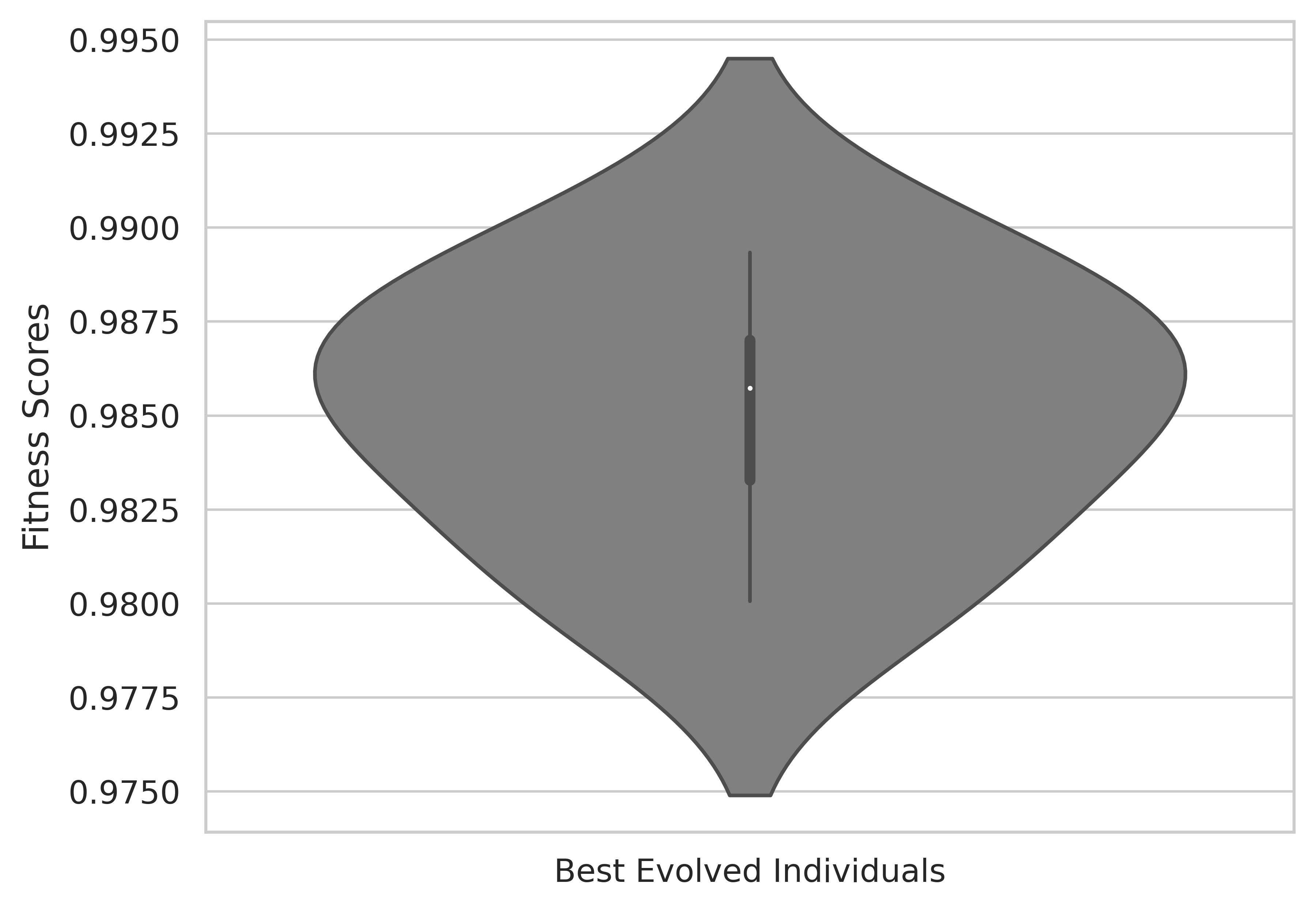}
    \label{fig:mnist_evo_violin}
}

\caption{Evolutionary analysis of SPENSER on the MNIST dataset over 200 generations. The results are averaged over 5 runs.}
\label{fig:mnist}
\end{figure}

\begin{figure}[htp]
\centering
\subfigure[Best Fitness]{
    \includegraphics[width=.31\textwidth]{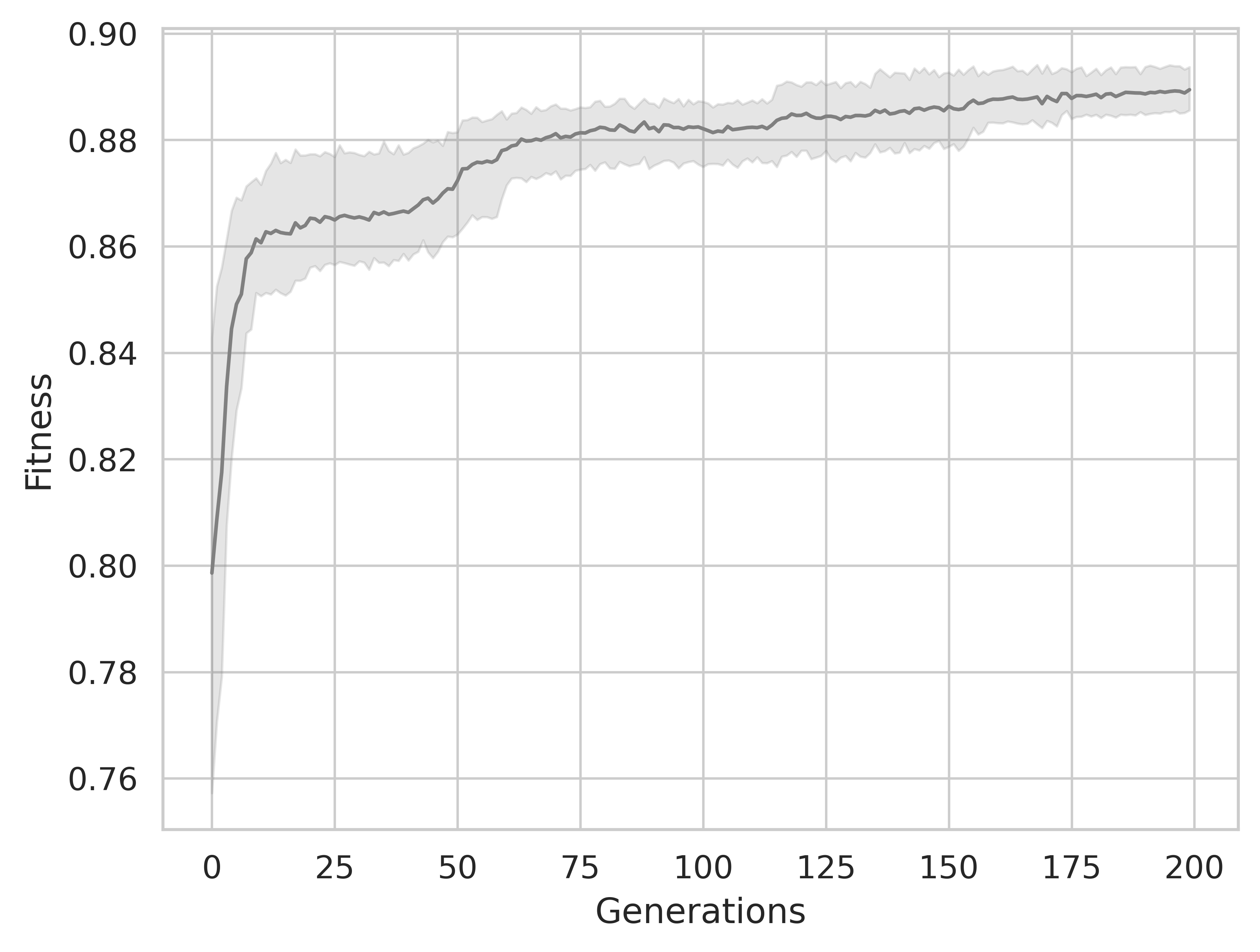}
    \label{fig:fashion_mnist_best}
    
}
\subfigure[Average Fitness]{
    \includegraphics[width=.31\textwidth]{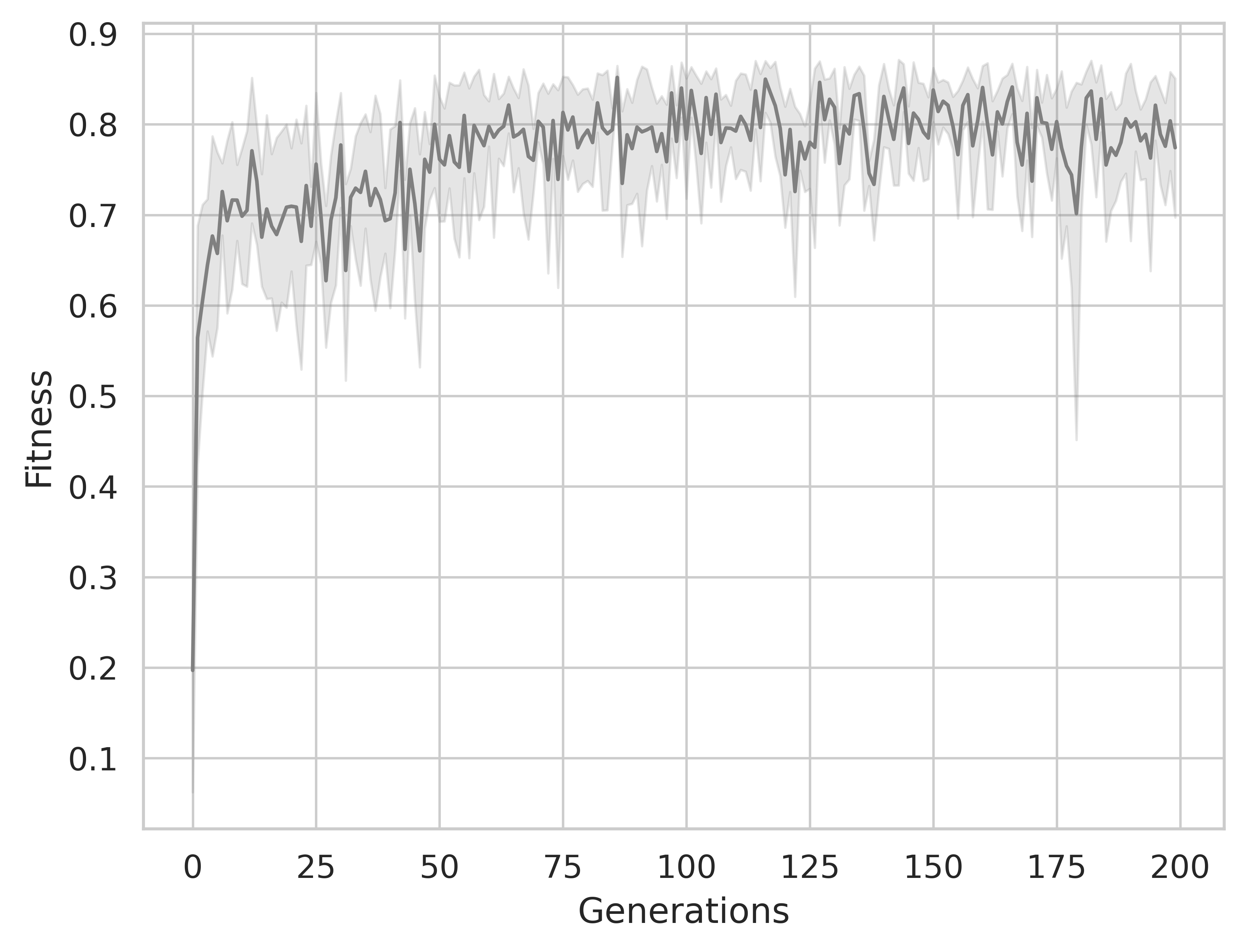}
    \label{fig:fashion_mnist_average}
}
%\\[\smallskipamount]
% \subfigure[New Best Individuals]{
%     \includegraphics[width=.31\textwidth]{figs/fashion_new_parent.png}
%     \label{fig:fashion_mnist_new}
% }
\subfigure[Fitness of best found individuals]{
    \includegraphics[width=.31\textwidth]{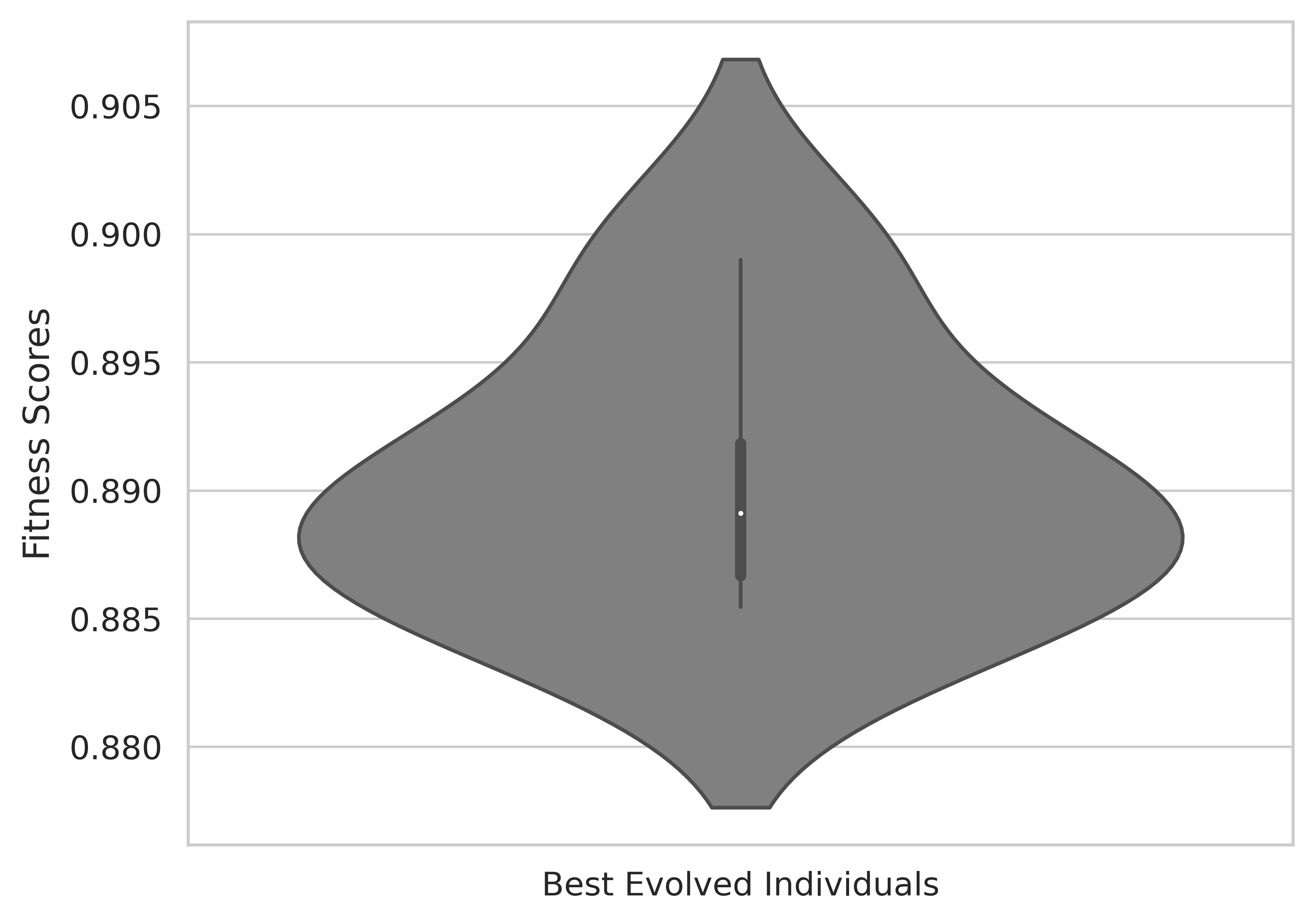}
    \label{fig:fashion_mnist_evo_violin}
}

\caption{Evolutionary analysis of SPENSER on the Fashion-MNIST dataset over 200 generations. The results are averaged over 5 runs.}
\label{fig:fashion_mnist}
\end{figure}

The evolutionary results are promising and show that SPENSER is able to generate increasingly better-performing individuals. Figures \ref{fig:mnist} and \ref{fig:fashion_mnist} display the evolution of the best fitness and the average fitness of the population across 200 generations, and a violin plot of the fitness of the best found individuals in the MNIST and Fashion-MNIST datasets respectively. The more notable aspects of the evolutionary search are the constant increase in best fitness and the diminishing variance over generations (Fig. \ref{fig:mnist_best}, \ref{fig:fashion_mnist_best}). These aspects showcase SPENSER's ability to uncover new and better individuals, and its consistency over different runs in generating better performing individuals. Furthermore, the average fitness of the population also increases, particularly in the Fashion-MNIST dataset (Fig. \ref{fig:fashion_mnist_average}), which demonstrates SPENSER's stability, as a random search would yield a constant average fitness. 

In order to understand if there are any notably better design choices for CSNNs, we summarized the best individuals' (both from MNIST and Fashion-MNIST) characteristics in Table \ref{tab:stats}. The most interesting result is the total absence of Average Pooling layers, as Kim et al. \cite{kim_neural_2022} had also stated that Average Pooling is not preferred for SNNs during their NAS and degraded performance. Furthermore, it is interesting to notice that the ATan surrogate gradient is preferred over the Fast-Sigmoid. The choice of Adam as a preferred optimizer is not surprising as it usually is the best-performing optimizer of the three.

\begin{table}[htp]
\caption{Network characteristics (percentage) for the best 10 individuals from MNIST and Fashion-MNIST.}
\begin{tabular}{llll}
\textbf{Layer Types}         &  & \textbf{Reset Mechanism} &  \\
Convolutional                                 & 35\%                                 & Subtract                                  & 63\%       \\
Average Pooling                               & 0\%                                  & Zero                                      & 37\%       \\
Max Pooling                                   & 19\%                                 &                                           &            \\
Dropout                                       & 11\%                                 & \textbf{Optimizers}      &            \\
Fully-Connected                               & 35\%                                 & Adam                                      & 70\%       \\
                                              &                                      & SGD                                       & 20\%       \\
\textbf{Surrogate Gradients} &                                      & RMSProp                                   & 10\%       \\
ATan                                          & 76\%                                 &                                           &            \\
Fast-Sigmoid                                  & 24\%                                 &                                           &           
\end{tabular}
\label{tab:stats}
\end{table}

%\vfill

%\vfill 

\subsection{Test Results}

\begin{figure}[h]
    \centering
    \includegraphics[width=7cm]{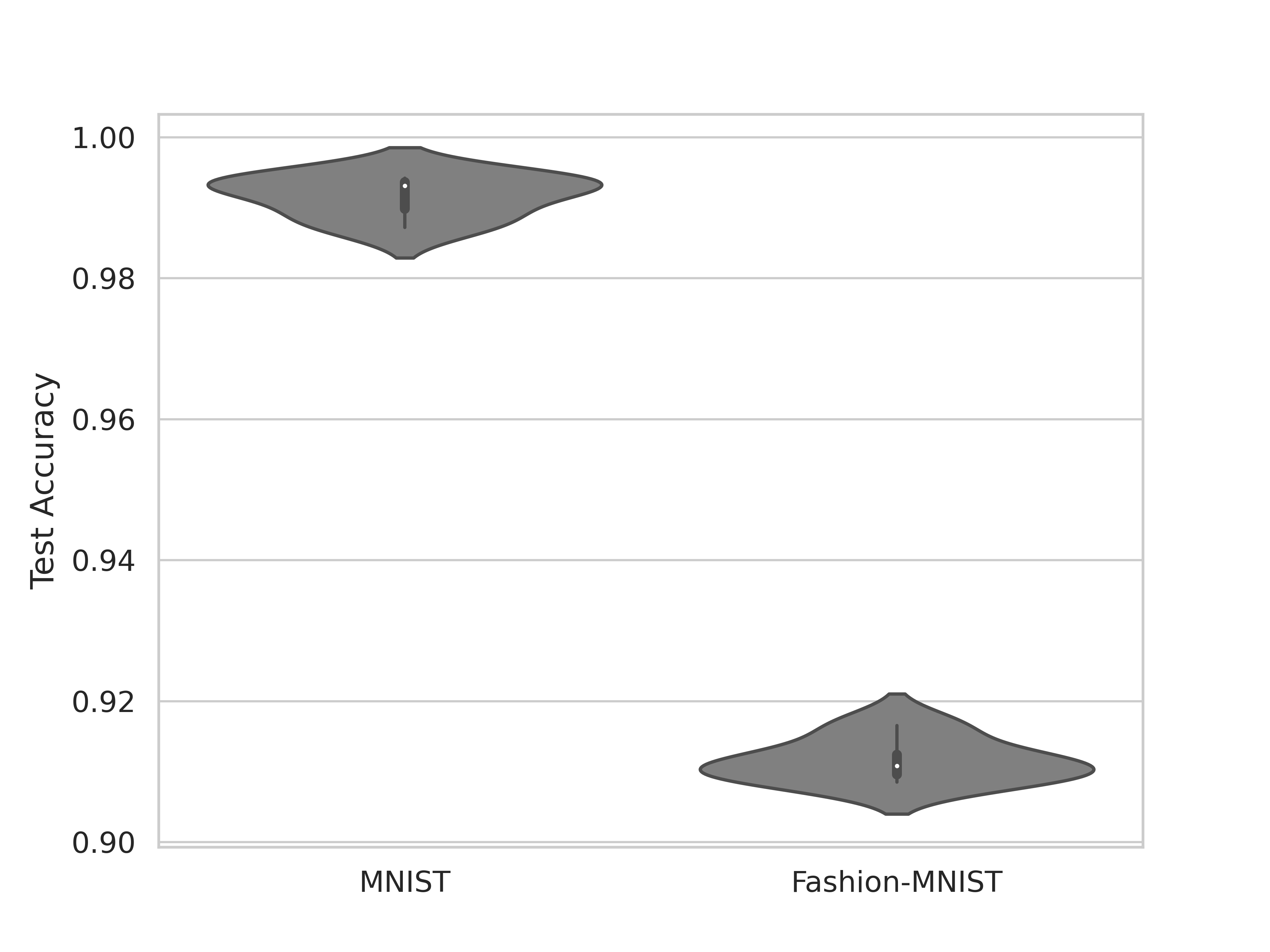}
    \caption{Violin plots of the Test accuracy of the best individuals after further training for 50 epochs.}
    \label{fig:test}
\end{figure}

After evolving for 200 generations, the best individuals were trained further for another 50 epochs (totaling 51 epochs) and evaluated on the Test set. Violin plots of the test accuracy on the MNIST and Fashion-MNIST datasets are displayed in Fig. \ref{fig:test}. Test results of different have small variations, showcasing SPENSER's robustness in generating high-performing networks.

We compared the best attained test accuracy with other works that also trained hand-tailored networks through spike based backpropagation. A comparison of test results is presented in Tab. \ref{tab:test_acc}. Albeit not surpassing the state-of-the-art, networks generated by SPENSER are head-to-head with the best-performing networks in the literature.

% \usepackage{multirow}
\begin{table}[h]
\centering
\caption{Test accuracy comparison of state of the art and our work.}
\begin{tabular}{|l|c|c|}
\hline
                                                         & MNIST   & Fashion-MNIST \\ \hline
Zhang et al. \cite{zhang2019spike}       & 99.62\% & 90.13\%       \\
Cheng et al. \cite{cheng2020lisnn}       & 99.50\% & 92.07\%       \\
Fang et al. \cite{fang2021incorporating} & 99.72\% & 94.38\%       \\
Jiang et al. \cite{jiang2023klif}        & 99.61\% & 94.35\%       \\
\textbf{SPENSER (ours)}                  & 99.42\% & 91.65\%       \\ \hline
\end{tabular}
\label{tab:test_acc}
\end{table}

\begin{figure}[h]
    \centering
    \includegraphics[width=7cm]{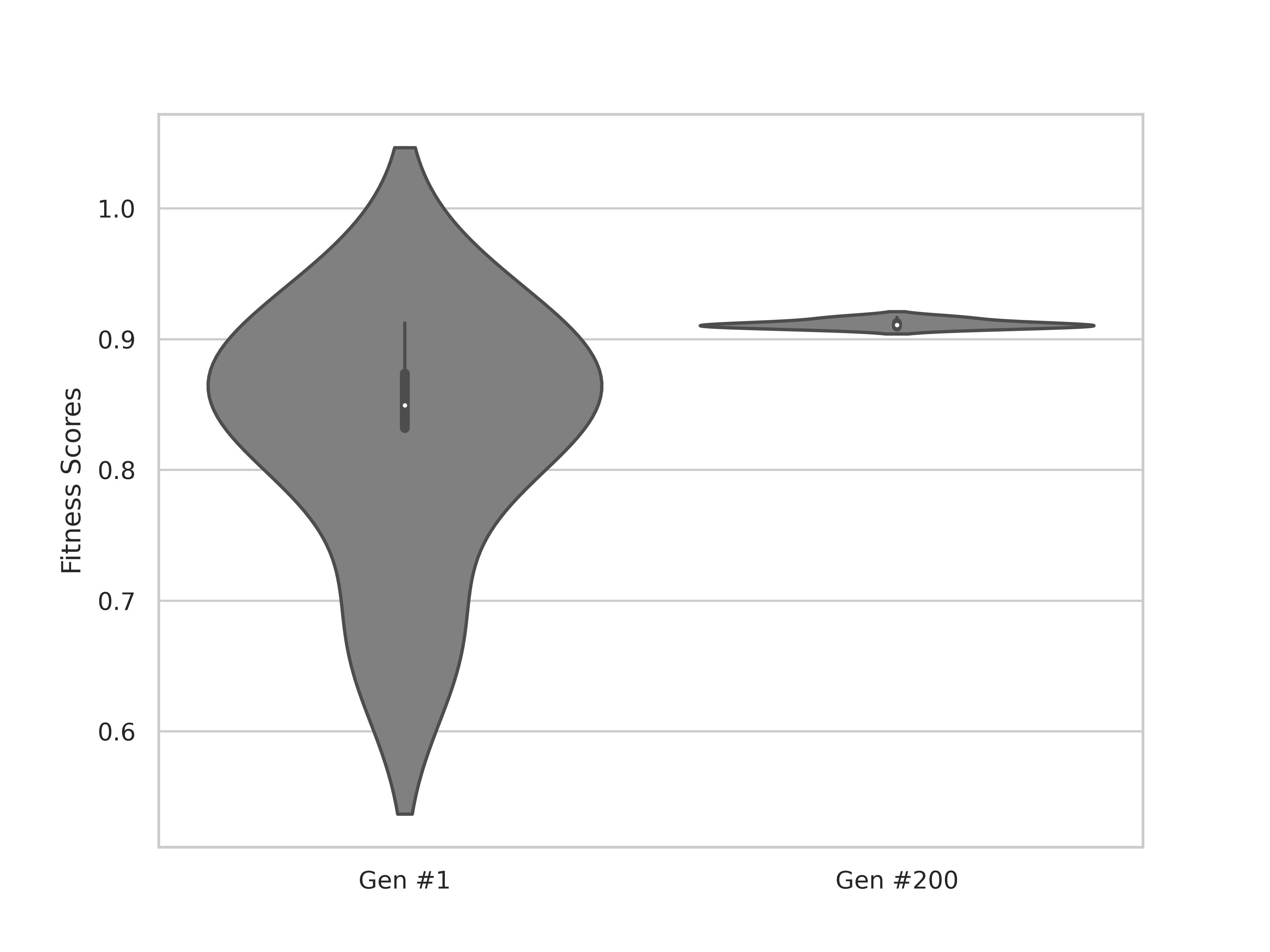}
    \caption{Test accuracy on Fashion-MNIST for the best individuals from Generation 1 and Generation 200, after 50 epochs of training.}
    \label{fig:fashion_0_200}
\end{figure}

In order to validate our choice of one epoch training for fitness assessment, we also trained the best networks found in the first generation of each run for another 50 epochs and tested their performance on the Test set. Fig. \ref{fig:fashion_0_200} displays violin plots for the test accuracy of the best individuals from generation 1 and generation 200. It is clear that the networks' performance is dependent on the architecture rather than training epochs and that the networks evolved by SPENSER perform better than random initialization. 

We hypothesize that a big limitation in this experimental study was the choice of the loss function's parameters, as it does not follow the literature's recommendations \cite{nunes2022spiking}. By setting the target firing rate of incorrect classes to 0\%, we might be suppressing output activity which is important to distinguish between closely distanced inputs. Furthermore, this experimental setup is sluggish, as training with BPTT is slower than in traditional ANNs and highly memory intensive. Kim et al. \cite{kim_neural_2022} have achieved impressive results without training the generated networks during the search phase, by estimating their future performance based on spike activation patterns across different data samples, and we believe this might be an important improvement to our framework. With faster experiments, we can focus on increasing diversity and coverage of the search space, so that SPENSER can yield better individuals.
\section{Final Remarks}\label{sec:conclusion}

In this paper we propose SPENSER, a NE framework to automatically design CSNNs. SPENSER is able to generate competitive performing networks for image classification at the level of the state of the art, without human parametrization of the network's architecture and parameters. SPENSER generated networks with competitive results, attaining 99.42\% accuracy on the MNIST \cite{mnist} and 91.65\% accuracy on the Fashion-MNIST \cite{fashion} datasets. Current limitations rely on the execution time, due to the computationally intensive BPTT learning algorithm and the memory requirements. Furthermore, we believe the configuration of the loss function played a role in suppressing output activity and potentially decreasing accuracy.

\subsection{Future Work}

In the future, we plan on:
\begin{itemize}
    \item Experiment with different loss functions / encode the loss function as an evolvable macro parameter;
    \item Perform a more in-depth study of the preferred choices during evolution and observable patterns in the best-performing individuals. This could be relevant in uncovering novel optimal architectures and parameters;
    \item Experiment with different learning algorithms.
    \item Implement skip connections and back connections.
    \item Apply regularisation methods to prevent vanishing and exploding gradients.
    
\end{itemize}

\begin{acks}
This research was supported by the Portuguese Recovery and Resilience Plan (PRR) through project C645008882-00000055, Center for Responsible AI, by the FCT - Foundation for Science and Technology, I.P./MCTES through national funds (PIDDAC), within the scope of CISUC R\&D Unit - UIDB/00326/2020 or project code UIDP/00326/2020. The first author is partially funded by FCT - Foundation for Science and Technology, Portugal, under the grant 2022.11314.BD.
\end{acks}

\bibliographystyle{ACM-Reference-Format}
\bibliography{references}

\end{document}